%% file: main.tex
\documentclass[10pt,twocolumn,letterpaper]{article}

\usepackage[pagenumbers]{cvpr} % To force page numbers, e.g. for an arXiv version
\usepackage{threeparttable}

\usepackage{multirow}
\usepackage[accsupp]{axessibility}
\usepackage{listings}\usepackage{xcolor}
\definecolor{cvprblue}{rgb}{0.21,0.49,0.74}
\usepackage[pagebackref,breaklinks,colorlinks,citecolor=cvprblue]{hyperref}
\usepackage[normalem]{ulem}
\usepackage{threeparttable}
\usepackage{xurl}
\usepackage{xcolor}
\definecolor{forestgreen}{rgb}{0.13, 0.55, 0.13}
\definecolor{darkorange}{rgb}{1.0, 0.55, 0.0}
\usepackage{colortbl}
\usepackage[normalem]{ulem}
\definecolor{mygray}{gray}{.9}
\definecolor{mypink}{rgb}{.99,.91,.95}
\definecolor{mycyan}{cmyk}{.3,0,0,0}
\definecolor{cvprblue}{rgb}{0.21,0.49,0.74}
\definecolor{grayblue}{rgb}{0.7, 0.75, 0.71}

\usepackage[pagebackref,breaklinks,colorlinks,citecolor=cvprblue]{hyperref}
\usepackage{multirow}
\usepackage{listings}\usepackage{xcolor}

\definecolor{codegreen}{rgb}{0,0.6,0}
\definecolor{codegray}{rgb}{0.5,0.5,0.5}
\definecolor{codepurple}{rgb}{0.58,0,0.82}
\definecolor{backcolour}{rgb}{0.95,0.95,0.92}

\lstdefinestyle{mystyle}{
  backgroundcolor=\color{backcolour}, commentstyle=\color{codegreen},
  keywordstyle=\color{magenta},
  numberstyle=\tiny\color{codegray},
  stringstyle=\color{codepurple},
  basicstyle=\ttfamily\footnotesize,
  breakatwhitespace=false,         
  breaklines=true,                 
  captionpos=b,                    
  keepspaces=true,                 
  numbers=left,                    
  numbersep=5pt,                  
  showspaces=false,                
  showstringspaces=false,
  showtabs=false,                  
  tabsize=2
}

\usepackage{xcolor}

\definecolor{codegreen}{rgb}{0,0.6,0}
\definecolor{codegray}{rgb}{0.5,0.5,0.5}
\definecolor{codepurple}{rgb}{0.58,0,0.82}
\definecolor{backcolour}{rgb}{0.95,0.95,0.92}

\lstdefinestyle{mystyle}{
  backgroundcolor=\color{backcolour}, commentstyle=\color{codegreen},
  keywordstyle=\color{magenta},
  numberstyle=\tiny\color{codegray},
  stringstyle=\color{codepurple},
  basicstyle=\ttfamily\footnotesize,
  breakatwhitespace=false,         
  breaklines=true,                 
  captionpos=b,                    
  keepspaces=true,                 
  numbers=left,                    
  numbersep=5pt,                  
  showspaces=false,                
  showstringspaces=false,
  showtabs=false,                  
  tabsize=2
}
\def\paperID{4744} % *** Enter the Paper ID here
\def\confName{CVPR}
\def\confYear{2024}

\title{Dynamic Prompt Optimizing for Text-to-Image Generation}

\author{Wenyi Mo\textsuperscript{1,2}, 
Tianyu Zhang\textsuperscript{3}, 
Yalong Bai\textsuperscript{3}, 
Bing Su\textsuperscript{1,2\thanks{Corresponding Authors.}}, 
Ji-Rong Wen\textsuperscript{1,2} and Qing Yang\textsuperscript{3}\\ 
\textsuperscript{1}Gaoling School of Artificial Intelligence, Renmin University of China \\
\textsuperscript{2}Beijing Key Laboratory of Big Data Management and Analysis Methods \\
\textsuperscript{3}Du Xiaoman Technology}

\begin{document}
\maketitle

\input{sec/0_abstract}

\input{sec/1_intro}

\input{sec/2_related}

\input{sec/3_method}

\input{sec/4_experiment}

\section{Conclusion}
\label{sec:conclusion}
In this paper, we propose PAE, a novel method for automatically editing prompts to improve the quality of images generated by a pre-trained text-to-image model. Unlike existing methods that require heuristic human engineering of prompts, PAE automatically edits input prompts and provides more flexible and fine-grained control. Experimental evaluations demonstrate the effectiveness and efficiency of PAE, which exhibits strong generalization abilities and performs well on both in-domain and out-of-domain data.

\noindent\textbf{Acknowledgment} This work was supported in part by the National Natural Science Foundation of China No. 62376277, Beijing Outstanding Young Scientist Program NO. BJJWZYJH012019100020098, and Public Computing Cloud, Renmin University of China.

{
    \small
    \bibliographystyle{ieeenat_fullname}
    \bibliography{main}
}

% WARNING: do not forget to delete the supplementary pages from your submission 

\input{sec/X_suppl}

\end{document}

%% file: sec/0_abstract.tex
\begin{abstract}

Text-to-image generative models, specifically those based on diffusion models like Imagen and Stable Diffusion, have made substantial advancements. Recently, there has been a surge of interest in the delicate refinement of text prompts.
Users assign weights or alter the injection time steps of certain words in the text prompts to improve the quality of generated images. 
However, the success of fine-control prompts depends on the accuracy of the text prompts and the careful selection of weights and time steps, which requires significant manual intervention. 
To address this, we introduce the \textbf{P}rompt \textbf{A}uto-\textbf{E}diting (PAE) method. 
Besides refining the original prompts for image generation, we further employ an online reinforcement learning strategy to explore the weights and injection time steps of each word, leading to the dynamic fine-control prompts. 
The reward function during training encourages the model to consider aesthetic score, semantic consistency, and user preferences. 
Experimental results demonstrate that our proposed method effectively improves the original prompts, generating visually more appealing images while maintaining semantic alignment. {Code is available at  \href{https://github.com/Mowenyii/PAE}{this https URL}.}

\end{abstract}

%% file: sec/1_intro.tex
\section{Introduction}
\label{introduction}

\begin{figure*}[h!]
	\vskip -0.08in
	\begin{center}
		\centering
		{\includegraphics[width=2.\columnwidth]{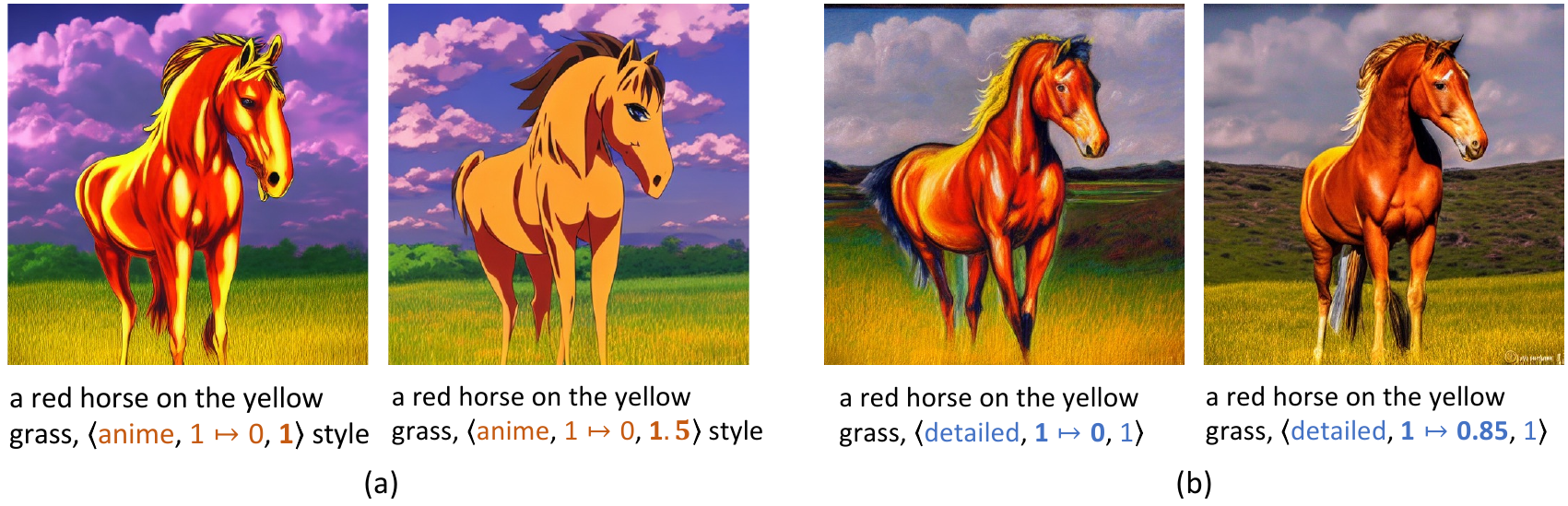}}
		\vskip -0.1in
		\caption{Generation results with the same seed using dynamic fine-control prompt (one plain token is extended into a triple of $\left\langle \text{token}, \text{effect range}, \text{weight} \right\rangle$). %with different weights or time steps. 
  It can be seen that (a) increasing the weight of \textcolor[rgb]{0.773,0.353,0.07}{anime} to \textbf{1.5} can amplify the sense of anime; (b) applying the word \textcolor[rgb]{0.267,0.447,0.769}{detailed} in the first \textbf{15\%} denoising timesteps can generate more natural texture details than applying it in all timesteps.}
		\label{fig:motivation}
	\end{center}
	\vskip -0.3in
\end{figure*}

Text-to-image generative models take a user-provided text to generate images matching the description~\cite{Bao2022AllAW,Nichol2022GLIDETP,Saharia2022PhotorealisticTD,Rombach2022HighResolutionIS}. 
The input text is called a prompt since it prompts the generative models to follow the user's instructions. However, it has been reported that recent text-to-image models are sensitive to prompts~\cite{prompt-DEHOUCHE2023e16757,prompt-Liu2022DesignGF,prompt-oppenlaender2022taxonomy}.
The organization of the input prompts plays a crucial role in determining the quality and relevance of the generated images. Interestingly, even when two prompts convey identical meanings, different expressions of these prompts may yield vastly different image interpretations.
Therefore, it is crucial to craft appropriate prompts that convey the user's intended ideas and establish clear communication with the generative model.

For a given pre-trained text-to-image generative model, it is unclear which type of prompt is the most suitable.  
Consequently, users heavily rely on heuristic engineering methods~\cite{Pavlichenko2022BestPF} by repeatedly running the generative model with modified prompt candidates in search of an optimal one.   
They append modifier words to enhance the art style or emphasize the image quality. 
These hand-crafted heuristics need to be implemented separately for each design intention and generative model, resulting in a costly, time-consuming, and labor-intensive trial-and-error process. 
Although there are learning-based methods~\cite{hao2022optimizing,DBLP:journals/corr/abs-2305-11317} that aim to enhance the quality of image generation results by rephasing or appending modifiers to user-input prompts, these methods lack control over the extent to which the added modifier words influence the image generation process. 

It is a common practice to assign varying levels of importance to specific words in the design of text prompt\footnote{\url{https://github.com/AUTOMATIC1111/stable-diffusion-webui/wiki/Features\#attentionemphasis}}.  This technique allows for more precise control over the generation process, as illustrated in \cref{fig:motivation}~(a).  
Another notable characteristic of the diffusion model is the multi-step denoising process. 
This multi-step design allows us to use different prompts at different time steps, thus achieving better results.
By precisely adjusting the effect time range of modifier words during this process, a significant enhancement of the visual aesthetics of the generated image can be achieved, as shown in \cref{fig:motivation}~(b). Therefore, to achieve more precise and detailed control over various aspects of the generated image, we propose a novel prompt format called the Dynamic Fine-control Prompt (DF-Prompt). It consists of  several triples of tokens, effect ranges, and importance levels. Traditional hand-crafted heuristic prompt engineering approaches struggle to handle such intricate and granular adjustments. Hence, it is necessary to develop an automated method for providing fine-grained optimization of prompts.

In this study, we propose a method called \textbf{P}rompt \textbf{A}uto-\textbf{E}diting (PAE). The primary aim of PAE is to optimize user-provided plain prompts to DF-Prompts for generating high-quality images. This optimization process is achieved through reinforcement learning. PAE involves a two-stage training process. In the first training stage of PAE, we introduce an automated method to overcome the dependency on manually constructed training samples. We define a confidence score to automatically filter publicly available prompt-image data. It ensures that the selected images are both visually pleasing and semantically consistent with the corresponding text. We then use this filtered dataset to fine-tune a pre-trained language model. The result is a tailored model that can enhance a given prompt with suitable modifiers. 
The second stage of PAE is based on the tailored model. 
We use online reinforcement learning tasks to encourage the model to explore better combinations of prompts and extra parameters, \textit{i.e.}, the effect range and weight of each modifier. 
To support this, we build a multidimensional reward system that takes into account factors such as aesthetic ratings, consistency between image and text semantics, and user preferences. Through the above process, PAE can automatically find the appropriate dynamic fine-grained prompt tokens. To demonstrate the effectiveness of our approach, we apply PAE to optimize text prompts from several public datasets, including Lexica.art\footnote{\url{https://lexica.art/}}, DiffusionDB~\cite{Wang2022DiffusionDBAL}, and COCO~\cite{cocodataset}. The experimental results show that our method can greatly improve human preference and aesthetic score while maintaining semantic consistency between the generated images and the original prompts. The contributions are as follows.
\begin{itemize}[leftmargin=20pt, itemsep=0pt, topsep=0pt]
\item Dynamic fine-control prompt editing framework: We introduce a framework that enhances prompt editing flexibility.  By integrating the effect range and weight of modifier tokens into a reinforcement learning framework, we enable fine-grained control and precise adjustments in image generation.

\item Effective results: Our method's effectiveness is thoroughly validated through experiments on several datasets.  The results show that our approach improves image aesthetics, ensures semantic consistency between prompts and generated images, and aligns more closely with human preferences.

\item Insightful findings: Our research reveals that artist names and texture-related modifiers enhance the artistic quality of generated images, while preserving the original semantics. It is more effective to introduce these terms in the latter half, rather than the initial half of the diffusion process. Assigning a lower weight to complex terms promotes a more balanced image generation. These findings hold significant implications for creative work and future research.

\end{itemize}

%% file: sec/2_related.tex
\begin{figure*}[ht!]
	\vskip -0.08in
	
		\centering
		\includegraphics[width=1.93\columnwidth]{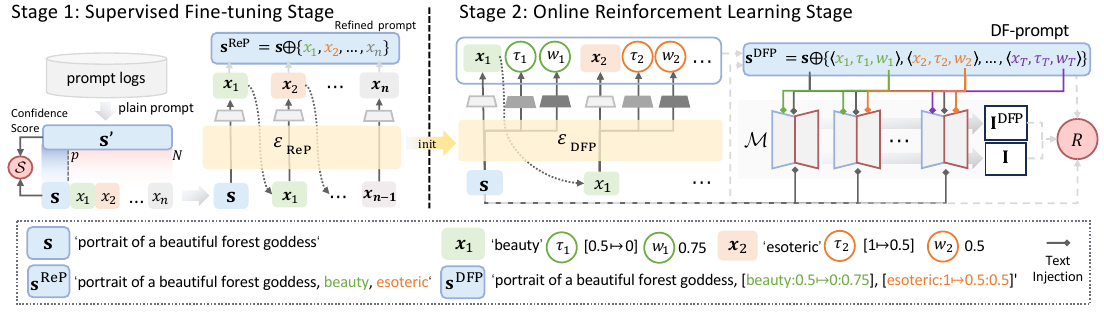}
		%\vskip -0.1in
		\caption{
The training process of PAE. \textbf{(Stage 1)} We select the training prompts based on a confidence score $\mathcal{S}$  as shown in~\cref{eq:confidence_score}, then fine-tune a pre-trained language model. The result is  $\mathcal{E}_\mathrm{ReP}$, a model that produces refined prompts. \textbf{(Stage 2)} We initialize the policy model $\mathcal{E}_\mathrm{DFP}$ using $\mathcal{E}_\mathrm{ReP}$. We add two linear headers to this model. These headers, along with the one predicting word tokens, use the same model's intermediate representation for their predictions. We then transform these predictions into DF-prompts. These DF-prompts modify the text injection mode of the diffusion model $\mathcal{M}$, which in turn affects the output images. During the online exploration, we use the original plain prompt $\mathbf{s}$, the optimized DF-prompt ${\mathbf{s}}^\mathrm{DFP}$, and their respective images $\mathbf{I}$ and ${\mathbf{I}}^\mathrm{DFP}$ to compute the reward $R$. Finally, we update the policy model by minimizing a loss function as defined in~\cref{eq:online_object}. 
}
	
	\vskip -0.2in
 \label{fig:overall}
\end{figure*}

\section{Related work}

\noindent\textbf{Content generation}\quad
 AI-generated content (AIGC)~\cite{DBLP:conf/icml/RameshPGGVRCS21,Saharia2022PhotorealisticTD,Rombach2022HighResolutionIS,Ramesh2022HierarchicalTI,DBLP:journals/corr/abs-2211-08332,peebles2023scalable,DBLP:journals/jmlr/ChowdheryNDBMRBCSGSSTMRBTSPRDHPBAI23,DBLP:journals/corr/abs-2302-13971} has made revolutionary progress in recent years, particularly in natural language processing. Large language models such as BERT~\cite{DBLP:journals/corr/abs-1810-04805}, GPT-1 to GPT-4~\cite{Radford2018ImprovingLU,Radford2019LanguageMA,Brown2020LanguageMA,OpenAI2023GPT4TR}, and ChatGPT\footnote{\url{https://chat.openai.com/}} have demonstrated exceptional text understanding and generation ability. Their advancements have greatly influenced the generation of text-to-image content. With the development of generative models~\cite{Esser2020TamingTF,SohlDickstein2015DeepUL,su2023intriguing,su2023unified} and multi-modal pre-training techniques~\cite{Radford2021LearningTV}, text-to-image generative models such as DALL·E 2~\cite{Ramesh2022HierarchicalTI}, Imagen~\cite{Saharia2022PhotorealisticTD}, Stable Diffusion~\cite{Rombach2022HighResolutionIS} and Versatile Diffusion~\cite{DBLP:journals/corr/abs-2211-08332} have showcased impressive performance in generating high-quality images. These breakthroughs have captured the attention of both academia and industry due to their potential impact on content production and applications in the open creative scene, \textit{etc}. In this paper, the proposed dynamic prompt editing framework utilizes a language generation model to assist text-to-image generation.

 \noindent\textbf{Text-to-image prompt collection and analysis}\quad In recent years, several studies have been conducted to explore the generative ability of text-to-image generative models. Some researchers collect prompt-image pairs from online communities or expert users~\cite{Wang2022DiffusionDBAL,DBLP:conf/www/XiePMJM23,Xu2023ImageRewardLA,DBLP:journals/corr/abs-2306-08141}. DiffusionDB~\cite{Wang2022DiffusionDBAL}, containing 2 million images, is collected from online public Stable Diffusion servers. It provides a valuable resource for researchers to study and improve the performance of text-to-image models. More recently, Xu \textit{et al.}~\cite{Xu2023ImageRewardLA} build an expert comparison dataset, including 137K prompt-image pairs from text-to-image models. These pairs are evaluated in terms of aesthetics, text-image alignment, toxicity, and biases. With the wealth of data, we aim to develop an automatic prompt editing method that can improve the performance of text-to-image models and generate high-quality images that satisfy users' demands.

\noindent\textbf{Prompt design}\quad
Text-to-image generative models~\cite{Crowson2022VQGANCLIPOD,DBLP:journals/jmlr/HoSCFNS22,DBLP:journals/corr/abs-2204-06125,Ramesh2022HierarchicalTI,Nichol2022GLIDETP,Saharia2022PhotorealisticTD,Rombach2022HighResolutionIS} are currently experiencing significant advancements, resulting in impressive visual effects from the generated images. However, these models only yield satisfactory images given appropriate input prompts, leading users to invest considerable time in modifying the prompts to ensure the generated images are aesthetically pleasing. In pursuit of higher-quality images, both researchers and online communities contribute creatively to prompt engineering for text-to-image generation~\cite{Oppenlaender2022TheCO,Pavlichenko2022BestPF,Wang2022DiffusionDBAL}. For instance, Pavlichenko \textit{et al.}~\cite{Pavlichenko2022BestPF} employ the genetic algorithm~\cite{Goldberg1988GeneticAI} to select a range of prompt keywords that enhance the quality of the images. Concurrently, Oppenlaender~\cite{prompt-oppenlaender2022taxonomy} utilizes auto-ethnographic research to understand the prompt design of online communities and categorizes existing prompt modifiers into six categories. Additionally, Liu \textit{et al.}~\cite{prompt-Liu2022DesignGF} collect over a thousand prompts for multiple group comparison experiments and proposes design guidelines for text-to-image prompt engineering. 
Recently, Hao \textit{et al.}~\cite{hao2022optimizing} propose a learning-based prompt optimization method using reinforcement learning. 
These approaches primarily focus on modifications to plain prompts and fail to achieve fine-control information injection. In this paper, we introduce a novel prompt editing framework to achieve fine-control prompt optimization. A reinforcement learning strategy is used to develop the capability of extending modifiers, adjusting weights, and adaptively fitting effect step ranges of the modifier tokens, with aesthetics, text-image semantic consistency, and human preferences serving as the reward.

%% file: sec/3_method.tex
\section{Method}
\label{sec:method}
In this section, we introduce the novel prompt format for diffusion-based text-to-image generative models. To achieve automated prompt editing, we design a two-stage training process, called Prompt Auto-Editing (PAE). PAE includes a supervised fine-tuning stage for refined prompt generation and an online reinforcement learning stage for dynamic fine-control prompt generation.

\subsection{Definitions of Dynamic Fine-control Prompt}
Given a pre-trained text-to-image generative model $\mathcal{M}$ and user input text $\mathbf{s}$, our goal is to produce a modified prompt $\mathbf{s}^m$ with fine-grained control so that the generated image, $\mathbf{I}^m \sim \mathcal{M}(\mathbf{s}^m)$, exhibits enhanced visual effects while remaining faithful to the semantics of the initial prompt $\mathbf{s}$. The modified prompt $\mathbf{s}^m$ contains the initial prompt $\mathbf{s}$ and a set of predicted modifiers $\mathbf{A}=\{x_1,\cdots,x_i,\cdots,x_n \}$, \textit{i.e.}, $\mathbf{s}^m =\mathbf{s} \oplus \mathbf{A}$. The $\oplus$ symbol indicates the append operation.

We hereby define a new prompt format that enriches the information of the initial prompt, named \textit{Dynamic Fine-Control Prompt} (DF-Prompt). Within this paradigm, each token $x_i$ of the modifier set $\mathbf{A}$ is coupled with an effect range $\tau_i$ and a specific weight $w_i$, resulting in a triple ${a}_i\!=\!\left\langle x_i,\tau_i,w_i\right\rangle$, where $w_i$ is a float number that weights the token embeddings for controlling the overall influences of token $x_i$ during image generating. The range $\tau_i\!=\![b_i\!\mapsto\!e_i]$ $(
1\!\ge\!b_i\!\ge\!e_i\!\ge\!0)$ is the normalized range that delineates the start and end steps during the iterative denoising process of the text-to-image model. 
We define the DF-Prompt token set is $\mathbf{A}^\mathrm{DFP}=\{\langle x_1,\tau_1,w_1\rangle,\cdots,\langle x_n,\tau_n,w_n\rangle\}$, and  the DF-Prompt is $\mathbf{s}^\mathrm{DFP}=\mathbf{s}\oplus\mathbf{A}^\mathrm{DFP}$. The essence of DF-Prompt lies in facilitating a more precise and controlled generation, ensuring the refined prompts are optimally structured for $\mathcal{M}$ to process. In order to facilitate demonstration and code implementation, we also define a plain-text format, where the triples are written within square brackets, [token:range:weight]. For instance, as shown in~\cref{fig:overall}, a DF-Prompt is written as ``portrait of a beautiful forest goddess, \text{[\textcolor{forestgreen}{beauty : $0.5\!\mapsto\!0$ : 0.75}], [\textcolor{orange}{esoteric : $1\!\mapsto\!0.5$ : 0.5}]}''.

\subsection{Overview of PAE}
{We formulate the prompt editing problem as a reinforcement learning task and propose a \textbf{P}rompt \textbf{A}uto-\textbf{E}diting method named PAE.} PAE enhances the user-provided prompt by adding modifiers in an auto-regressive manner while assigning corresponding effect ranges and weights. As illustrated in~\cref{fig:overall}, PAE operates in two distinct training stages. \textbf{Stage 1}: To enrich simple prompts, we fine-tune a pre-trained language model on a curated prompt-image dataset. The dataset is specifically selected based on a confidence score $\mathcal{S}$. The result of this stage is a refined prompt model $\mathcal{E}_\mathrm{ReP}$. \textbf{Stage 2}: This stage involves an online reinforcement learning process. We implement a policy model $\mathcal{E}_\mathrm{DFP}$ initialized from  $\mathcal{E}_\mathrm{ReP}$. The policy model interacts with the environment (the text-to-image model $\mathcal{M}$) through the current policy (the model-derived mapping from the input prompt to the dynamic fine-control prompt). A reward function is defined to evaluate the aesthetic appeal of the generated image, its semantic similarity to the input text, and its alignment with human preference. The policy model $\mathcal{E}_\mathrm{DFP}$ is then optimized based on a defined loss function.

\subsection{Finetuning for Plain Prompt Refinement}
\label{sec:collect_data}
In the first stage, we utilize selected data to fine-tune the \text{GPT-2}~\cite{Radford2019LanguageMA} model to get a plain prompt refining model ${\mathcal{E}}_\mathrm{ReP}$. The model ${\mathcal{E}}_\mathrm{ReP}$ predicts suffix modifiers one by one, and this process repeats until the model outputs the stop sign, \textit{i.e.}, \textless\textbar endoftext\textbar\textgreater. Given a prompt $\mathbf{s}$, we construct the refined prompt as ${\mathbf{s}}^\mathrm{ReP}\!=\! \mathbf{s} \oplus \mathbf{A}$, where~$\mathbf{A}~\sim~{\mathcal{E}}_\mathrm{ReP}(\mathbf{s})$.

\noindent\textbf{Data Selection.} Different from previous methods that depend on human-in-the-loop annotation datasets~\cite{Pavlichenko2022BestPF}, we collect training data from public text-image datasets and online communities.  Given the inconsistent quality of images in publicly available text-image pairs, not all prompts are suitable for model training. Therefore, we devise an automated process for data filtration and training sample construction. The rule for data filtration stipulates that \textit{only instances that demonstrate an improvement in aesthetics and maintain semantic relevance after the addition of modifiers are retained}. 
As depicted on the left of~\cref{fig:overall}, we start with a given prompt $\mathbf{s}'$ from publicly available prompt logs. 
The original prompt $\mathbf{s}'$ is split at a division point $p\in \{1,\cdots,N\} $. Here, $N$ represents the number of tokens in $\mathbf{s}'$. 
The text preceding the division point is considered to contain primary information, describing the main theme of the image; the text following the division point is regarded as secondary, providing supplementary suffixes as modifier words.
According to~\cite{Wang2022DiffusionDBAL}, we select the first comma in $\mathbf{s}'$ as the division point.
Following this, we obtain the short prompt $\mathbf{s}=\{s_1, ..., s_{p} \}$, which is the first $p$ tokens joined together. 
The remaining tokens form the modifier set $\mathbf{A}=\{x_1,\cdots,x_n|x_1=s_{p+1}, \cdots, x_n=s_{N} \}$.  
Lastly, we define a confidence score, $\mathcal{S}(\mathbf{s}, \mathbf{s}')$. Using this, we construct the training samples as follows:
\begin{equation}
\label{eq:confidence_score}
\begin{aligned}
    \mathbb{D} &= \left\{ \left\langle \mathbf{s}, \mathbf{A} \right\rangle \mid \mathcal{S}(\mathbf{s}', \mathbf{s})> 0 \right\}, \\
    \mathcal{S}(\mathbf{s}', \mathbf{s}) &= \mathbb{E}_{\mathbf{I}' \sim \mathcal{M}(\mathbf{s}'),\mathbf{I} \sim \mathcal{M}(\mathbf{s})} \big[  u\left(g_\mathrm{aes}(\mathbf{I}')-g_\mathrm{aes}(\mathbf{I})\right)\\
    &\quad \times u\left( g_\mathrm{CLIP}(\mathbf{s}, \mathbf{I}')-g_\mathrm{CLIP}(\mathbf{s}, \mathbf{I}) + \gamma \right) 
    \big],
\end{aligned}
\end{equation}
where $g_\mathrm{CLIP}$ measures the image-text relevance by using pre-trained CLIP model~\cite{Radford2021LearningTV} and $g_\mathrm{aes}$ returns the aesthetic score\footnote{\url{https://github.com/christophschuhmann/improved-aesthetic-predictor}}. The parameter \( \gamma \) acts as a tolerance constant. Additionally, \( u(z) \) represents a characteristic function that returns 1 if \( z > 0 \) and 0 otherwise.

We train the language model based on the training datasets $\mathbb{D}$ using teacher forcing methods~\cite{teacherforcing}, and perform a direct auto-regressive style negative log-likelihood loss on the next token:
\begin{equation}
\mathcal{L}_\mathrm{ReP}= -\mathbb{E}_{\left\langle\mathbf{s}, \mathbf{A} \right\rangle\sim \mathbb{D}} \left[\log {P(\mathbf{A}|\mathbf{s},\mathcal{E}}_\mathrm{ReP})\right].
\end{equation}
In this way, the trained model $\mathcal{E}_\mathrm{ReP}$ is proficient in handling brief prompt inputs, \textit{i.e.}, simple text describing the image theme, and predicting appropriate modifiers to formulate refined prompts $\mathbf{s}^\mathrm{ReP}$, thereby elevating the aesthetic quality of the generated image.

\subsection{RL for DF-Prompt Generation}

In the second training stage, we aim to explore better prompt configurations by specifying effect ranges and weights for additional modifier suffixes.

\noindent\textbf{{\textbf{Online reinforcement learning.}}}\quad We utilize PPO algorithm~\cite{Schulman2017ProximalPO}, a popular reinforcement learning method known for its effectiveness and stability. 
The aim is to maximize the expected cumulative reward over the training set $\mathbb{D}$. We add two head layers on ${\mathcal{E}}_\mathrm{ReP}$ to predict the effect range and weight corresponding to each token, and initialize the parameters of additional layers to output \( \tau_i = [1\mapsto0] \) and \( w_i = 1 \) for every token $x_i$. 
After that, ${\mathcal{E}}_\mathrm{ReP}$ is used to initialize a policy model $\mathcal{E}_\mathrm{DFP}$. During an episode of prompt optimization, we set the initial state as the initial text $\mathbf{s}=\{s_1, ..., s_{p} \}$. The action space is tripartite: word space $\mathcal{V}$, discrete time range space $\mathcal{T}=\{0.5\mapsto0,1\mapsto0,1\mapsto0.5\}$, and discrete weight space $\mathcal{W}=\{0.5,0.75,1,1.25,1.5\}$. 
At each step $t$ of online exploration, the model selects an action $a_t=\left\langle x_t,\tau_t,w_t | x_t \in \mathcal{V}, \tau_t\in \mathcal{T},  w_t\in\mathcal{W}\right\rangle$, in accordance with the policy model $a_t \sim  \mathcal{E}_\mathrm{DFP}(\mathbf{s}_{<t})$. To be consistent with the input format of the language model, we define the state at $t$-th step with tokens only, \textit{i.e.}, $\mathbf{s}_{<t}=\mathbf{s}\oplus\{x_1,x_2,\cdots,x_{t-1}\}$.

During training, the policy model $\mathcal{E}_\mathrm{DFP}$ interacts with the text-to-image model $\mathcal{M}$. We make adjustments to the text encoder module of the model, with the specific implementation details outlined in supplementary materials. 
These modifications allow for weighting individual tokens and customizing the effective time range during the denoising process.
The predicted action set $\mathbf{A}^\mathrm{DFP}=\{\langle x_1,\tau_1,w_1\rangle,\cdots,\langle x_T,\tau_T,w_T\rangle\}$ are used to generate images. 
Using the generated images, we compute the reward $R(\mathbf{s}, \mathbf{A}^\mathrm{DFP})$. We define a loss function $\mathcal{L}_\mathrm{DFP}$, which is used to optimize the policy model:
\begin{equation}
\label{eq:online_object}
\vspace{-2.5pt}
\mathcal{L}_\mathrm{DFP} = -\mathbb{E}_{\mathbf{s} \sim \mathbb{D}, \mathbf{A}^\mathrm{DFP} \sim \mathcal{E}_\mathrm{DFP}}\left[ R(\mathbf{s}, \mathbf{A}^\mathrm{DFP}) -\eta D_\mathrm{KL}\right],
\end{equation}
where $D_\mathrm{KL}$ computes the Kullback-Leibler divergence~\cite{KLD}. It serves as a regulation constraint to minimize differences between the output modifiers of the policy model $\mathcal{E}_\mathrm{DFP}$ and those of the initial model $\mathcal{E}_\mathrm{ReP}$~\cite{DBLP:conf/nips/Ouyang0JAWMZASR22}. We also use Gaussian distributions to supervise the effect range probability distribution and weight distribution predicted by $\mathcal{E}_\mathrm{DFP}$. More implementation details are in~\cref{sec:imp_detail}.

Another component in PPO is the value model. Its role is to estimate the expected cumulative reward from the current state, directed by the policy model's actions. Its optimization objective is to minimize the difference between the predicted and actual rewards. In the optimization process, the policy model and the value model are optimized alternately, so that they can promote each other to maximize the expected cumulative reward. We initialize the value model with $\mathcal{E}_\mathrm{ReP}$ and replace the initial linear layer with a regression head for better performance.

\noindent\textbf{{\textbf{Reward definition.}}}\quad We construct the reward $R(\mathbf{s}, \mathbf{A}^\mathrm{DFP})$ using CLIP Score, Aesthetic Score, and PickScore~\cite{kirstain2023pick}:
\begin{equation}\label{eq:reward}
\begin{split}
R(\mathbf{s}, \mathbf{A}^\mathrm{DFP})=&\mathbb{E}_{\mathbf{I} \sim \mathcal{M}(\mathbf{s}),\mathbf{I}^\mathrm{DFP} \sim \mathcal{M}(\mathbf{s}\oplus \mathbf{A}^\mathrm{DFP})}[ \\& \min \left( g_\mathrm{CLIP}\left(\mathbf{s}, \mathbf{I}^\mathrm{DFP}\right)-\zeta,0\right)\\&
+\min \left( g_{\mathrm{PKS}}\left(\mathbf{s}, \mathbf{I}^\mathrm{DFP} \right)-\kappa,0\right)
\\& 
+ \alpha \cdot\left(g_{\mathrm{aes}}(\mathbf{I}^\mathrm{DFP}) - \beta\cdot g_{\mathrm{aes}}({\mathbf{I}})\right)]. 
\end{split}
\end{equation}
where $g_\mathrm{PKS}$ denotes the learned human preference evaluation metric of PickScore. {The symbols \( \zeta \) and \( \kappa \) set minimum thresholds for CLIP score and PickScore contributions to the reward, while \( \alpha \) and \( \beta \) scale the Aes score's impact.}

%% file: sec/4_experiment.tex
\section{Experiments}
\label{sec:experiments}

\subsection{Experimental Setup}
\noindent\textbf{Data Collection.} The public text-image pair sources include Lexica.art and DiffusionDB~\cite{Wang2022DiffusionDBAL}. The NSFW images are recognized with an image classification model and removed from the training data. 
After that, we conduct data selection as described in~\cref{sec:collect_data}. Finally, we get about $450,000$ prompts. 
We randomly select 500 $\left\langle\mathbf{s}, \mathbf{s}'\right\rangle$ pairs from DiffusionDB for validation and extract $1,000$ prompts from Lexica.art and DiffusionDB respectively for evaluation. 
In particular, we also use $1,000$ prompts randomly selected from COCO~\cite{cocodataset} dataset for out-of-domain evaluation. 
The training set, validation set, and test set are independent of each other.

\noindent\textbf{Comparison to other methods.}
We compare the prompts edited with our method to four types of prompts: the short primary prompts $\mathbf{s}$, the original human-written prompts $\mathbf{s}'$, and the prompts generated from the same short prompt s by the pre-trained GPT-2~\cite{Radford2019LanguageMA} and Promptist~\cite{hao2022optimizing}. {
Human-written prompts are randomly chosen from user-provided prompt datasets like Lexica.art and DiffusionDB, while short prompts are the texts before the first commas.}

\begin{figure*}
	\vskip -0.08in
	\begin{center}
		\centering
            {\includegraphics[width=1.0\linewidth,page=1]{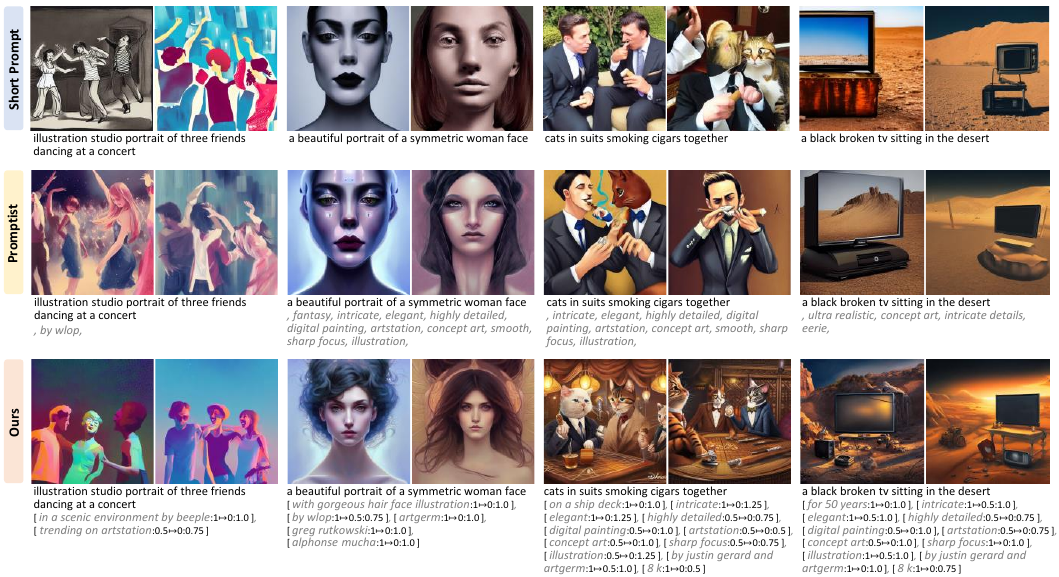}}
		\vskip -0.1in
		\caption{Generated images using Stable Diffusion v1.4 with short prompts, Promptist~\cite{hao2022optimizing}, and our method. In each column, the images are generated using the same random seed. Our method shows the ability to moderately expand the semantic content, such as ``in a scenic environment'', ``with gorgeous hair face illustration'', ``on a ship deck'' and ``for 50 years.'' These expansions stimulate users' imagination while enhancing the comprehensiveness and aesthetic quality of the image.}
		\label{fig:cmp_promptist}
	\end{center}
	\vskip -0.25in
\end{figure*}

\noindent\textbf{Metrics.}
We utilize four metrics to evaluate the results of edited prompts: Aesthetic score, CLIP score~\cite{Radford2021LearningTV}, PickScore~\cite{kirstain2023pick}, {CMMD score~\cite{DBLP:journals/corr/abs-2401-09603}.}
The Aesthetic score reflects the visual attractiveness of an image. Higher values indicate better visual quality. The CLIP score evaluates the alignment between the generated image and the prompt. PickScore is an automatic measurement standard used to comprehensively assess the visual quality and text alignment of images. Larger values indicate a greater consistency between the generated image and human preferences. {CMMD offers a more accurate and consistent measure of image quality by not assuming a normal distribution of data and being efficient with sample sizes. Lower CMMD values indicate more realistic images.}
In our evaluation, we report the Aesthetic scores of the corresponding images, CLIP scores between the short prompt and the images generated by the edited prompt. For PickScore, We report the relative pairwise comparisons $\mathbb{E}[g_{\mathrm{PKS}}(\mathbf{s}, {\mathbf{I}}^m)\ge g_{\mathrm{PKS}}(\mathbf{s}, {\mathbf{I}})]$ between the edited prompt $\mathbf{s}^m$ and the short prompt $\mathbf{s}$. 
{We report CMMD between the generated images and the real images corresponding to the prompts in the COCO dataset.}

\subsection{Implementation Details}
\label{sec:imp_detail}
For the processes of data collection, model training, and evaluation, we use Stable Diffusion v1.4~\cite{Rombach2022HighResolutionIS} with the UniPC solver~\cite{Zhao2023UniPCAU}, and set the inference time steps to 10. 

\begin{figure}[t]
	% \vskip -0.08in
	\begin{center}
		\centering

{\includegraphics[width=1.\linewidth,page=2]{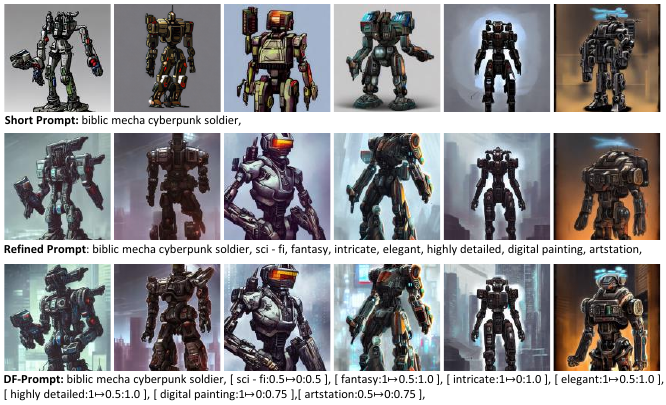}}
		\vskip -0.1in
		\caption{Our method generate the DF-Prompt, which corresponds to the generated images with more detailed textures and a richer background for a better visual effect than the refined prompt. The images are generated using the same random seed in each column.
  }
		\label{fig:cmp_plain}
	\end{center}
	\vskip -0.3in
\end{figure}

\begin{figure*}[t]
	\vskip -0.1in
		\centering
\includegraphics[width=1.9\columnwidth]{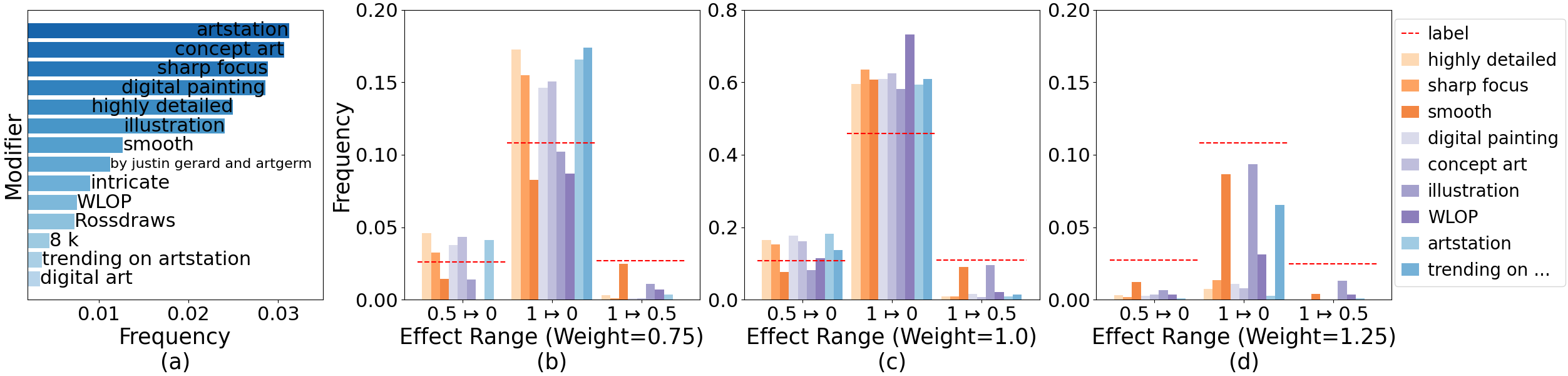}
		\vskip -0.1in
		\caption{(a) The 15 most frequently generated modifiers. (b$\sim$d) The frequency of different combinations of settings.}
	\vskip -0.1in
 \label{fig:freq}
\end{figure*}

\noindent\textbf{\textbf{Supervised fine-tuning}.} Empirically, we find that when training with the default settings for both effect range and weight ($\tau_i = [1\mapsto0]$ and $w_i = 1$) as a one-point distribution, the policy model is prone to overfitting to this settings. To address this, we apply a strategy similar to Label Smoothing~\cite{Szegedy2015RethinkingTI} in the first stage to enhance the model's learning process. This strategy involves sampling discrete values from Gaussian distributions. The means of these distributions are consistent with the values of the default settings for effect range and weight, and they share a uniform variance of $\sigma$. The frequency of different joint settings is shown in the dotted line marked by ``label'' in~\cref{fig:freq}~(b$\sim$d). This introduction of random sampling from Gaussian distributions aims to diversify the training signal in the first stage, thereby enabling better generalization in second stages. For the model structure of $\mathcal{E}_\mathrm{ReP}$, we load the pre-trained GPT-2 Medium~\cite{Radford2019LanguageMA} weights and add two linear heads directly to approximate the distributions. We can use the distributions predicted by these heads to supervise the effect range probability distribution and weight distribution predicted by $\mathcal{E}_\mathrm{DFP}$. We train the model for 50k steps, using a batch size of 64 and a learning rate of $5\times10^{-5}$, with the Adam optimizer. The block size is 256. To avoid the model learning fixed patterns, we introduce variability by randomly altering the case of the prompt's first letter and replacing commas with periods at a 50\% probability. {
In our implementation, phrases separated by commas share the same effect range and weight, calculated using the mode of the range and weight among these phrases.}

\noindent\textbf{\textbf{Online reinforcement learning}.} In our experiment, we follow the approach by Hao \textit{et al.}~\cite{hao2022optimizing} to set $\zeta=0.28$ in the reward function. The stability of the rewards is crucial in our process. To ensure this, we calculate the reward by generating two images per prompt. We train both the policy and the value models for 3,000 episodes, each with a batch size of 32. For optimization, we set the learning rate to $5\times10^{-5}$ and employ the Adam optimizer for both models. We adjust the Adam optimizer's hyper-parameters, setting $\beta_1$ to 0.9 and $\beta_2$ to 0.95. The KL coefficient $\eta$ is 0.02. To save memory, we use a simplified version of the PPO algorithm that processes one PPO epoch per batch.

\subsection{Evaluation and Analysis}

\noindent\textbf{\textbf{Qualitative analysis.}} As shown in~\cref{fig:cmp_promptist}, based on the short prompts, PAE adds texture-related terms like ``highly detailed'', the artist's name ``justin gerard and artgerm'', and some highly aesthetically related words ``elegant'', ``artstation'' to enhance the aesthetic quality of the generated images. As shown in~\cref{fig:cmp_plain}, the DF-prompt generated by our method can provide finer control than the refined prompt.

\begin{table}[t]
\vskip -0.1in
\small
\centering
\resizebox{\linewidth}{!}{\begin{tabular}{lccccc}
\toprule
Method        &PickScore ($\uparrow$)   & CLIP Score ($\uparrow$)    & Aes Score ($\uparrow$) \\\midrule
Short Prompt &  - & {0.28}&   5.58 \\\midrule
GPT-2  & 47.9\% &0.25 &       5.38 \\
Human & \uline{72.5\%}& \uline{0.26}&         {6.07}\\
Promptist  & 68.4\%& \textbf{0.27}&     \uline{6.11}\\
PAE (Ours) &  \textbf{73.9\%}  &\uline{0.26}&    \textbf{6.12} \\\bottomrule
\end{tabular}}\vskip -0.1in
\caption{Quantitative comparison on Lexica.art. }\label{tab:Lexica.art}
\vskip -0.2in
\end{table}

\begin{table}[h]
\vskip -0.1in
\small
    \centering
    \begin{tabular}{lc}
    \toprule
         Method & CMMD ($\downarrow$)  \\
      \hline
    Promptist & 1.147 \\
    PAE (Ours)  & \textbf{1.125}\\
      \bottomrule
    \end{tabular}
    \vskip -0.1in
    \caption{{Quantitative comparison using the CMMD metric.}}
    \label{tab:cmmd}
  \vskip -0.15in
\end{table}

\noindent\textbf{\textbf{Quantitative comparison.}} We evaluate PAE on two in-domain datasets: Lexica.art and DiffusionDB. As shown in~\cref{tab:Lexica.art} and~\cref{tab:DiffusionDB}, the results show that PAE surpasses other methods in terms of Aesthetic Score, and it achieves a human preference Pick Score that closely mirrors the human-written prompt. This suggests that PAE aligns well with human aesthetic preferences. Additionally, we evaluate PAE on the out-of-domain dataset COCO. As shown in~\cref{tab:COCO}, PAE outperforms other methods in terms of Pick Score. This consistent performance across various datasets demonstrates the robustness and versatility of the PAE method. {Furthermore, as shown in~\cref{tab:cmmd}, PAE outperforms Promptist, as it indicates lower CMMD scores~\cite{DBLP:journals/corr/abs-2401-09603}. This shows that the prompts edited by our method generate images of superior quality and enhanced realism.}

\begin{table}[t]
\small
\small
\centering
\vskip -0.1in
\resizebox{\linewidth}{!}{\begin{tabular}{lccccc}
\toprule
Method        &PickScore ($\uparrow$)   & CLIP Score ($\uparrow$)    & Aes Score ($\uparrow$) \\\midrule
Short Prompt  & -       & {0.28}&5.58\\ \midrule
GPT-2           &48.1\%  &0.25&  5.40 \\
Human         &\textbf{70.5\%} &\uline{0.26}& 5.84 \\
Promptist &62.3\% &\textbf{0.27}& \uline{6.06}\\
PAE (Ours) & \uline{64.4\%} & \uline{0.26} &\textbf{6.07} \\\bottomrule
\end{tabular}} \vskip -0.1in
\caption{Quantitative comparison on DiffusionDB. }  
\vskip -0.16in
\label{tab:DiffusionDB}
\end{table}

\begin{table}[t]
\small
\centering
\resizebox{\linewidth}{!}{
\begin{tabular}{lccccc}\toprule
Method        &PickScore ($\uparrow$)   & CLIP Score ($\uparrow$)    & Aes Score ($\uparrow$) \\\midrule
Short Prompt &  - & {0.27 }& 5.37\\\midrule
GPT-2         & 51.2\%  &0.25 & 5.24\\
Promptist &\uline{53.4\%} &0.25 &\textbf{6.15} \\
PAE (Ours)& \textbf{53.8\%}  &0.25 &\uline{6.09}
\\\bottomrule
\end{tabular}}\vskip -0.1in
\caption{Quantitative comparison on COCO.}\label{tab:COCO}
\vskip -0.25in
\end{table}

\noindent\textbf{\textbf{Statistical analysis of text.}} We apply our method to 3,500 prompts, gathering DF-prompt tokens from the policy model. The top 15 frequently generated modifiers are displayed in \cref{fig:freq}~(a). They mainly pertain to art trends such as ``artstation'', artist names like ``WLOP'', art styles and types such as ``digital painting'' and ``illustration'', and texture-related terms like ``highly detailed'' and ``smooth''. These modifiers subtly boost the artistic vibe without significantly altering the prompt's semantics. In \cref{fig:freq}~(b$\sim$d), the red dotted lines indicate the frequency of the label case as detailed in~\cref{sec:imp_detail}. We observe several phenomena and attempt to interpret them: \textbf{1)} In (c), most terms mentioned above appear more frequently than the label case under the $1\mapsto0$ and $0.5\mapsto0$ settings. This suggests that these effect ranges yield higher rewards during training when the weight is 1.0, hence the policy model leans towards selecting them. \textbf{2)} Also in (c), the $0.5\mapsto0$ setting outperforms the $1\mapsto0.5$ setting. This suggests that injecting texture-related terms and art styles (except ``smooth'' and ``illustration'') into the final 50\% of diffusion time steps is more effective than in the first 50\%. This latter half of the diffusion time steps is typically when image details and structure start to form. Hence, it's optimal to introduce texture-related terms and art styles at this stage, as they can directly impact the image's details and structure. Conversely, introducing these elements in the initial 50\% of the diffusion time steps may not significantly influence the final image, as these elements could be overwhelmed by subsequent diffusion steps when the image is still relatively unstructured. \textbf{3)} Comparing the $1\mapsto0$ setting in (b) and that in (d), the setting with weight~$=0.75$ occurs more frequently than weight~$=1.25$. By assigning a lower weight (0.75), the prompt effectively instructs the generative model to pay less attention to these tokens. This could lead the model to consider all tokens more evenly when generating images, resulting in a more balanced and potentially superior outcome. Furthermore, these elements (like ``digital painting'', ``concept art'', ``artstation'', \textit{etc}.) are inherently complex and can be interpreted in various ways. If the model focuses excessively on the tokens (due to the higher weight of 1.25), it might struggle to generate coherent images due to these concepts' complexity and ambiguity. Note that the aforementioned observations merely reflect the trends, different prompts may have different optimal choices, which is why our method is necessary.

\subsection{Ablation Study} We conduct ablation experiments on the DiffusionDB validation dataset to examine the effects of different data settings, training settings, and prompt types. 

\noindent\textbf{\textbf{Data Settings.}}
The main parameters associated with the training data are a variance $\sigma$ in~\cref{sec:imp_detail} and a tolerance constant $\gamma$ in~\cref{eq:confidence_score}. As shown in~\cref{tab:abl_sigma_gamma},
the setting of $\sigma=0.5,\,\gamma=0.01$ obtains the highest aesthetic score, so we choose it as the parameter setting for other experiments.

\begin{table}[!ht]
\small
    \centering

    \vskip -0.1in
        \begin{tabular}{ccc}\toprule
       Data Settings  & CLIP Score ($\uparrow$)& Aes Score ($\uparrow$)  \\ \midrule
        $\sigma=0.5,\,\gamma=0.00$  & 0.26 & 6.01 \\ 
        $\sigma=0.5,\,\gamma=0.01$  & 0.26 & \textbf{6.03}\\ 
        $\sigma=1.0,\,\gamma=0.00$ & 0.26 &5.95\\
        $\sigma=1.0,\,\gamma=0.01$ & 0.26& 5.94\\\bottomrule 
    \end{tabular}\vskip -0.1in
        \caption{Ablation experiments on hyperparameters of the validation set. We validate the results of the first-stage model $\mathcal{E}_\mathrm{ReP}$ at 50k steps on the DiffusionDB Validation set.}\label{tab:abl_sigma_gamma}
        \vskip -0.1in
\end{table}

\begin{table}[t]
\small
    \centering
    \resizebox{\linewidth}{!}{
    \begin{threeparttable}
    \begin{tabular}{ccccc}\toprule
   &Reward Settings    &Pick* ($\uparrow$) & CLIP ($\uparrow$)& Aes ($\uparrow$)  \\ \midrule
     (1) &  $\alpha=1,\,\beta=0,\,\kappa=16$ & 53.8\% & 0.26 & 6.01\\
        & $\alpha=1,\,\beta=0,\,\kappa=18$  & \textbf{58.0\%} & 0.26 & 6.04\\
        & $\alpha=1,\,\beta=0,\,\kappa=20$ & 56.4\% & 0.26 & \textbf{6.05}\\\midrule

    (2)    & $\alpha=1,\,\beta=1,\,\kappa=16$ & 3.8\% & \textbf{0.28} & 5.56\\
        & $\alpha=1,\,\beta=1,\,\kappa=18$ & 9.6\% & \textbf{0.28} & 5.54 \\ 
        & $\alpha=1,\,\beta=1,\,\kappa=20$ & 5.2\% & \textbf{0.28} & 5.56\\\midrule

      (3)  & $\alpha=5,\,\beta=1,\,\kappa=18$ & 52.0\% & 0.26 & 5.97\\
         & $\alpha=10,\,\beta=1,\,\kappa=18$ & 57.0\% & 0.26 & 5.93\\
        \bottomrule 
    \end{tabular}
        \begin{tablenotes}
\scriptsize
\item * To highlight the disparity, we report the measure $\mathbb{E}[g_{\mathrm{PKS}}(\mathbf{s}, {\mathbf{I}}^m) > g_{\mathrm{PKS}}(\mathbf{s}, {\mathbf{I}})]$.
\end{tablenotes}
\end{threeparttable}}\vskip -0.1in
        \caption{Ablation experiments on different parameters of reward. The second stage model $\mathcal{E}_\mathrm{DFP}$ is trained for 1,000 episodes. }\label{tab:abl_alpha}
        \vskip -0.1in
\end{table}
\noindent\textbf{\textbf{Training Settings.}}
In our method, the reward is primarily influenced by three main parameters: $\alpha$, $\beta$, and $\kappa$, as outlined in~\cref{eq:reward}. In \cref{tab:abl_alpha}, we observe that when $\kappa=18$, a higher PickScore is achieved, while the CLIP score and Aes Score remain relatively consistent compared to other values of $\kappa$. Comparisons between (1) and (2), setting $\beta=1$ results in a significant increase in the CLIP score, but leads to a decrease in both the aesthetic score and PickScore, compared to when $\beta=0$. Furthermore, in comparing (2) and (3), we find that an increase in $\alpha$ boosts both the latter scores, albeit at the cost of the CLIP score. Given that our task is primarily aimed at enhancing human preferences and aesthetics without causing significant semantic deviations, we choose $\alpha=1,\,\beta=0,\,\kappa=18$ for other experiments. We also demonstrate the improvement brought by the second stage of training. In \cref{tab:abl_init}, compared with $\mathcal{E}_\mathrm{ReP}$, the policy model $\mathcal{E}_\mathrm{DFP}$ can bring comprehensive improvement.

\begin{figure}[h!]
	\vskip -0.15in
		\centering
\includegraphics[width=1.0\columnwidth]{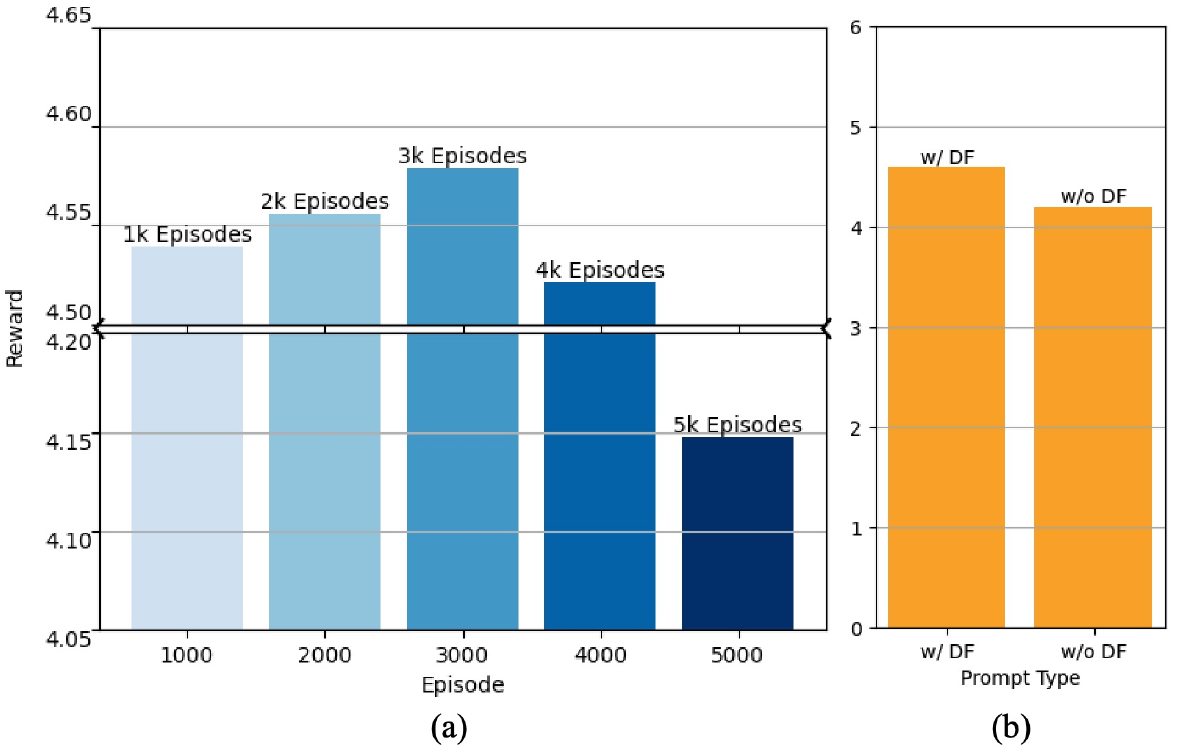}
		\vskip -0.13in
		\caption{(a) The relationship between episode and reward. (b) Ablation experiments with different prompt types.
}
	\vskip -0.25in
 \label{fig:episode_bar}
\end{figure}

\noindent\textbf{\textbf{Ablation experiments on different episodes.}}
As shown in~\cref{fig:episode_bar}~(a), the policy model achieves its peak reward after 3,000 episodes of training. Consequently, we adopt 3,000 episodes as the standard setting for other experiments. 

\noindent\textbf{\textbf{DF-prompt format.}}
As shown in~\cref{fig:episode_bar}~(b), when other settings remain the same, the reward increases with the output of the policy model using the DF-Prompt format instead of the plain prompt format. This indicates that compared to plain prompts, DF-Prompts enhance the aesthetic appeal of the generated images. They also strengthen the alignment between the image and the prompt, making the image more in line with human preferences.

\begin{table}[t]
% \vskip -0.05in

\vskip -0.05in
\small
\centering
% \vskip -0.05in
\resizebox{\linewidth}{!}{
\begin{tabular}{lcccccc}\toprule
Method        &PickScore* ($\uparrow$)   & CLIP Score ($\uparrow$)    & Aes Score ($\uparrow$) & Reward ($\uparrow$)\\\midrule
$\mathcal{E}_\mathrm{ReP}$&  53.8\% & \textbf{0.26}&6.03 & 4.49\\
$\mathcal{E}_\mathrm{DFP}$& \textbf{57.8\%} &\textbf{0.26}& \textbf{6.07} & \textbf{4.58}  \\\bottomrule
\end{tabular}} 
\vskip -0.1in
\caption{Comparison between the initial model $\mathcal{E}_\mathrm{ReP}$ and the second stage model $\mathcal{E}_\mathrm{DFP}$ trained over 3,000 episodes.}\label{tab:abl_init}
     \vskip -0.2in
\end{table}

%% file: sec/X_suppl.tex
\clearpage

\setcounter{page}{1}
\maketitlesupplementary

In this appendix, we provide additional information and materials to complement our research, including training samples, more qualitative results, additional experimental details, and discussion.

\section*{A. Examples of training data}
\begin{figure*}
	% \vskip -0.1in
	\begin{center}
		\centering
{\includegraphics[width=2.\columnwidth]{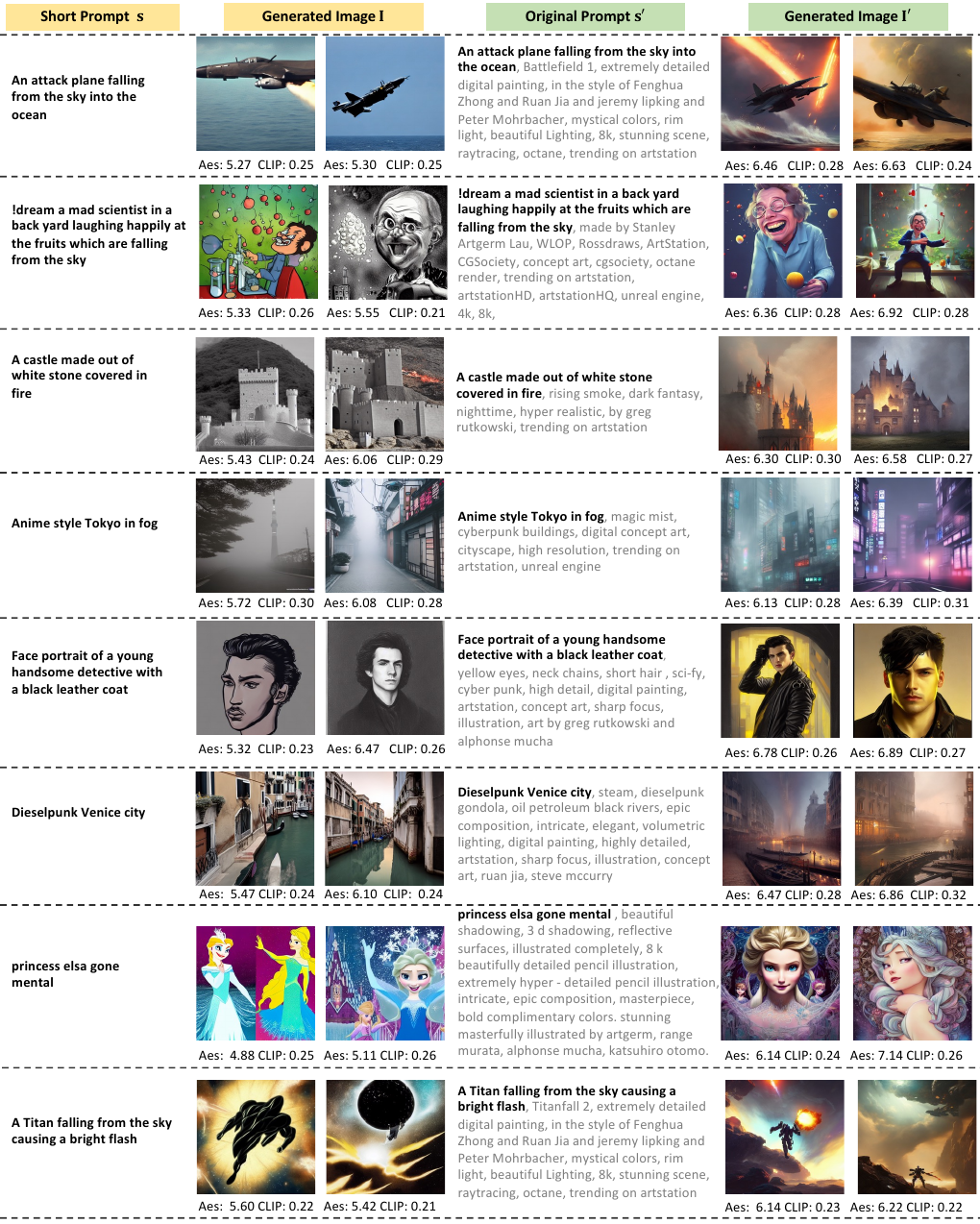}}
		% \vskip -0.1in
		\caption{Some examples of the training data.
  }
		\label{fig:train_example}
	\end{center}
	% \vskip -0.3in
\end{figure*}

We utilize a diverse range of text-image pairs sourced from public datasets and online communities. As shown in \cref{fig:train_example}, we present some prompts that are included in our training data. These prompts have undergone filtration and construction following the automated process described in \cref{sec:collect_data}. The short prompts $\mathbf{s}$ primarily describe the subject matter of the images, while the modifiers (highlighted in gray) provide additional details and enhance the aesthetic appeal of the images. In the figure, the term ``Aes'' denotes the aesthetic score, and ``CLIP'' quantifies the semantic relevance of the generated image to the short prompt. We can see that the generated images $\mathbf{I}'$ corresponding to the original prompt $\mathbf{s}'$ are more visually effective than the generated images $\mathbf{I}$ corresponding to the short prompt $\mathbf{s}$.

\begin{table}[h]
\centering
    \begin{tabular}{lcc}\toprule
    Lexica.art               & Aes  & CLIP \\\midrule
    Short Prompt&  5.58 & 0.28 \\ \hline %\cline{1-1}
 \rowcolor{blue!10}{ + ``artstation''}       & 5.83      & 0.26       \\ %\cline{1-1}
 \rowcolor{purple!10}      { + ``concept art''}      & 5.68      & 0.30        \\
 \rowcolor{purple!10}      { + ``digital painting''} & 5.79      & 0.30        \\%\cline{1-1}
    \rowcolor{orange!10}   { + ``sharp focus''}      & 5.60       & 0.28    \\
  \rowcolor{orange!10}   { + ``highly detailed''}  & 5.64      & 0.29     \\\bottomrule
    \end{tabular}\vskip -0.05in
      \caption{The effect of different words on generating images.} \label{tab:word}
\end{table}

\begin{table*}[t]
% \vskip -0.1in
\small
    \centering
\begin{tabular}{c|cc|cc|cc}
\toprule
   \multirow{2}{*}{Method }      & \multicolumn{2}{c}{Training} \vrule& \multicolumn{2}{c}{Inference (per prompt)}\vrule & \multicolumn{2}{c}{T2I Pipeline (per image)}\\\cline{2-7}%\cline{7-8}
   & Stage 1& Stage 2 & Ours & Promptist & Vanilla SD & Dynamic SD\\ \midrule
 GPU Times &   18 hours   & 3 days &  0.73s & 0.69s &  5.64s & 5.71s \\\bottomrule
\end{tabular} \vskip -0.1in
\caption{Experiment on an A800 (80GB) GPU.}  \label{tab:time}
% \vskip -0.30in
\end{table*}

\section*{B. More detailed statistical analysis}
\begin{figure}[h]
  \centering
  \includegraphics[width=1\linewidth]{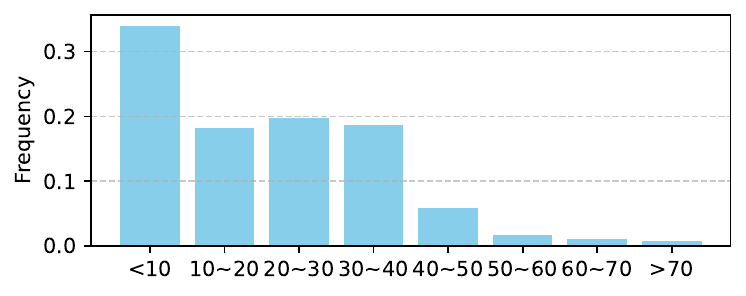}
  \vskip -0.1in
  \caption{Frequency of the number of predicted word tokens.}
  \label{fig:bar_token_num}
\end{figure}

\cref{fig:bar_token_num} indicates a predominance of shorter token sequences in model predictions, implying that adding a few modifiers can significantly enhance an image's visual appeal without altering the original prompt's meaning. 
\cref{fig:freq}~(b-d) show frequently generated modifiers, most of which are trends, styles, and texture terms. 
We also conduct experiments to analyze word impact. 
As shown in \cref{tab:word}, ``artstation'' boosts Aesthetic scores at the cost of text-image similarity, whereas styles and texture modifiers slightly increase Aesthetic scores while preserving alignment.

\begin{figure*}
	% \vskip -0.08in
	\begin{center}
		\centering
{\includegraphics[width=2\columnwidth]{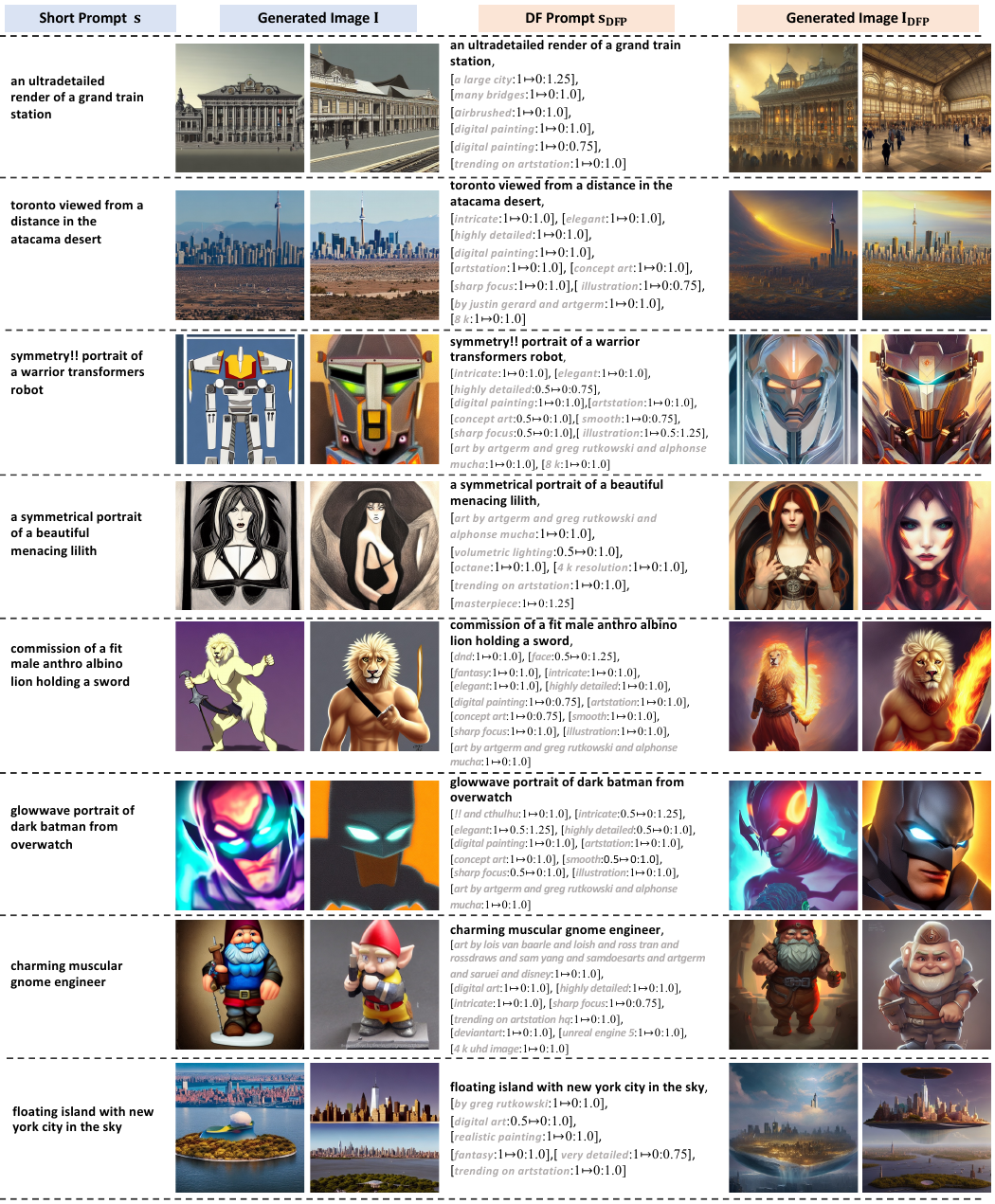}}
% {\includegraphics[width=0.3\linewidth,page=2]{figure/train_example.pdf}}
		% \vskip -0.1in
		\caption{More examples of the generated images.
  }
		\label{fig:sp_lexica}
	\end{center}
	% \vskip -0.3in
\end{figure*}

\begin{figure*}
	\vskip -0.08in
	\begin{center}
		\centering
{\includegraphics[width=2\columnwidth]{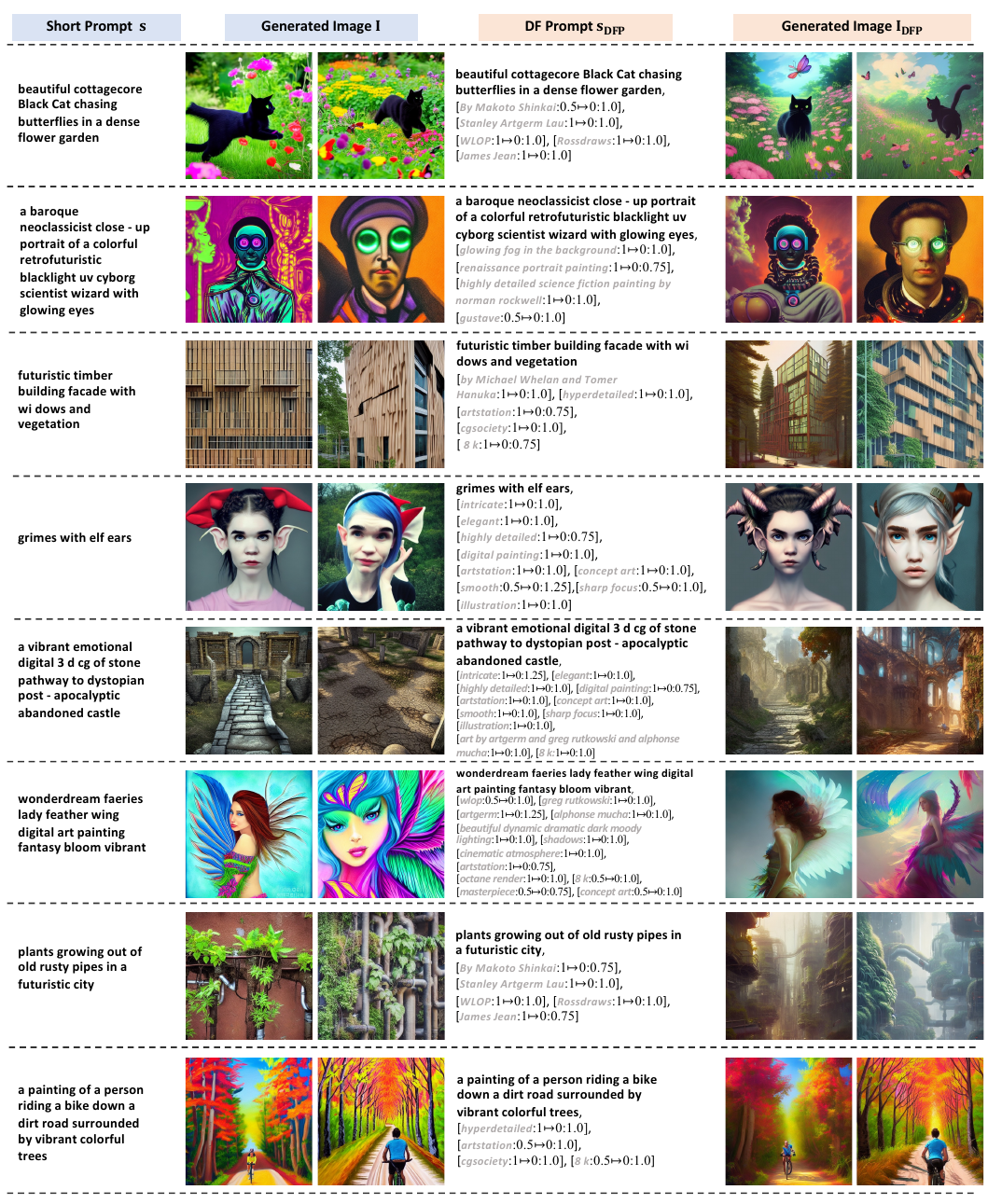}}
% {\includegraphics[width=0.3\linewidth,page=2]{figure/train_example.pdf}}
		\vskip -0.1in
		\caption{More examples of the generated images.
  }
		\label{fig:sp_diffusiondb}
	\end{center}
	\vskip -0.3in
\end{figure*}

\begin{figure*}
	% \vskip -0.1in
	\begin{center}
		\centering
		{\includegraphics[width=2.\columnwidth]{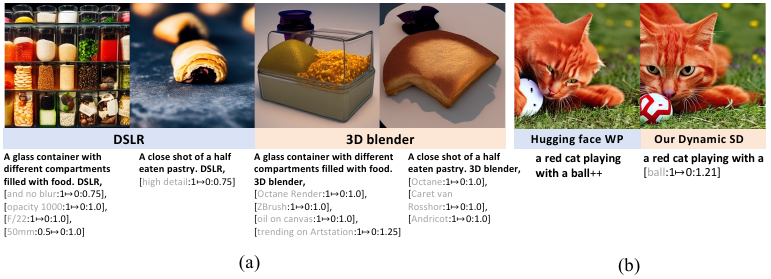}}
		% \vskip -0.13in
		\caption{(a) Examples of practicability. (b) Weight methods.}
		\label{fig:DSLR}
	\end{center}
	% \vskip -0.22in
\end{figure*}

\begin{figure*}
	% \vskip -0.1in
	\begin{center}
		\centering		{\includegraphics[width=2\columnwidth]{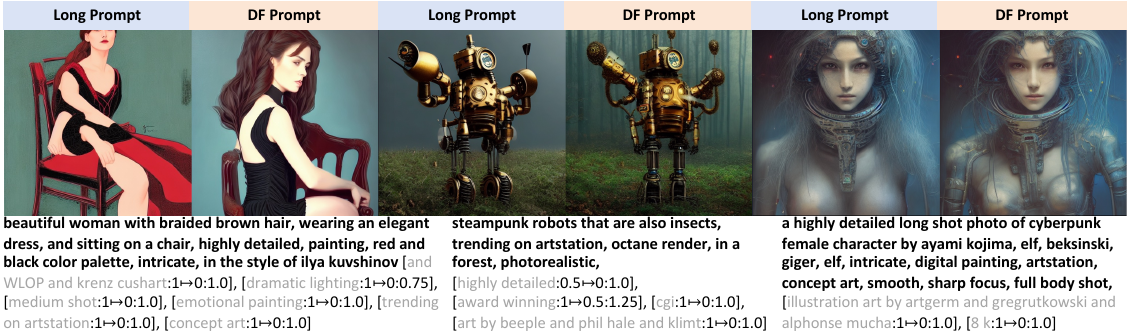}}
		% \vskip -0.12in
		\caption{The input long prompts are in bold.}
		\label{fig:long_prompt}
	\end{center}
	% \vskip -0.22in
\end{figure*}

\section*{C. Enhanced text encoder for DF-prompt}
\label{sec:text_encoder} In Stable Diffusion, the text encoder is modified to achieve fine-grained control over the generated effects. These modifications involve two key aspects:
\begin{itemize}
\item We introduce weights for each word embedding, representing the impact of a word or phrase on the resulting image. To accomplish this, we apply a weighting operation to each word's embedding by multiplying it with a specific weight. Subsequently, we normalize the entire set of text embeddings, ensuring that the overall mean value remains consistent with the original text embeddings. This normalization step is crucial for maintaining numerical stability. {Our technique yields results similar to the existing prompt weighting method (\cref{fig:DSLR}(b)) but having dynamic time-range control. 
The pseudo-code for weighting tokens is below. }
\captionsetup[lstlisting]{skip=.01pt}
\begin{lstlisting}[language=Python,caption=Python pseudo code for weighting tokens.]
# Given text_embs:[77x768], weights:[77,]
previous_mean = text_embs.mean() # float
text_embs *= weights
current_mean = text_embs.mean()
text_embs *= previous_mean / current_mean 
\end{lstlisting}

\item The injection time steps are regulated using a dictionary. This dictionary maps each word or phrase to a designated time step, which determines when to initiate and conclude the injection of that specific word or phrase during the image generation process. By manipulating the time steps in the dictionary, precise control over the duration of different concepts within the generated image can be achieved.
\end{itemize}
These modifications empower the text encoder to exert more precise control over the effects within the Stable Diffusion framework.  As a result, more personalized and user-specific image-generation outcomes can be attained.

\begin{table}[h]
\centering
    \begin{tabular}{cc}\toprule
     Method ( +``DSLR'') & 
      FID ($\downarrow$) \\ \midrule
      Promptist   &  70.80\\
      PAE (Ours)&\textbf{69.84} \\\bottomrule
    \end{tabular} \vskip -0.05in
    \caption{{Quantitative comparison of image quality between our method and Promptist, measured using the FID metric.}}
    \label{tab:fid}
\end{table}

\section*{D. More experimental details}
For the evaluation process, we use a maximum new token length of 75 for all evaluated models. We use a temperature of 0.9 during the evaluation and apply a top-k sampling strategy with a k-value of 200. To ensure consistency, we use the same seed in all quantitative evaluation experiments.

\section*{E. More qualitative results}
In this section, we present more qualitative results, as depicted in \cref{fig:sp_lexica,fig:sp_diffusiondb}. We compare the images $\mathbf{I}^\mathrm{DFP}$ generated using DF-Prompts with the images $\mathbf{I}$ generated using the short prompts. For example, in~\cref{fig:sp_lexica}, we observe that the images corresponding to DF-Prompts, $\mathbf{I}^\mathrm{DFP}$, exhibit more vibrant details and aesthetically pleasing color combinations compared to the images $\mathbf{I}$ generated from the short prompts. Some specific examples include ``symmetry!! portrait of a warrior transformers robot'', ``a symmetrical portrait of a beautiful menacing lilith'', ``commission of a fit male anthro albino lion holding a sword'' and ``glowwave portrait of dark batman from overwatch''. We ensure fairness and consistency by generating the columns corresponding to $\mathbf{I}$ and $\mathbf{I}^\mathrm{DFP}$ using the same seed.

{
Empirical evidence shows that our method not only creates aesthetically pleasing images but also caters precisely to user queries, such as achieving photorealism with ``DSLR'' or creating 3D-rendered effects with ``3D blender'' in user prompts (\cref{fig:DSLR}(a)). Our method shows adaptability when integrating detailed modifiers like ``DSLR'' and achieves competitive Frechet Inception Distance (FID)~\citep{DBLP:conf/nips/HeuselRUNH17} (\cref{tab:fid}). This adaptability is critical in practical applications.}

{The time cost of each stage is shown in~\cref{tab:time}. As for inference, the average time is marginally higher than that of Promptist (+0.04 s). 
Moreover, our Dynamic Stable Diffusion (Dynamic SD) method is slightly slower than the Vanilla SD method, but the difference is minimal.}

{

\section*{F. Discussion}

{The significant enhancement in image quality and text alignment observed in \cref{fig:cmp_promptist} of the main paper for the case ``cats in suits smoking cigars together'' can be attributed to our model's reward mechanism. Specifically, we incorporate the Aes score to encourage actions that improve aesthetic features and the CLIP score to ensure semantic coherence. Additionally, our reward function introduces the PickScore, which allows for more diverse prompt modifiers and ultimately leads to improved image quality. In \cref{fig:cmp_promptist}, the inclusion of new semantic elements like ``on a ship deck'' alongside other modifiers contributes significantly to the visual appeal of the generated output.}

{Differences among PAE, Promptist, \textit{hugging face weighting prompt method} (WP)\footnote{\url{https://huggingface.co/docs/diffusers/using-diffusers/weighted\_prompts}}  lie in that Promptist focuses solely on prompt expansion, while WP manipulates the likelihood of certain phrases appearing in images by artificially setting their weights. PAE, on the other hand, innovatively introduces dynamic prompts, and dynamically adjusts the weights of different phrases during various stages of image denoising, thus achieving more granular control over the image generation process. Additionally, PAE introduces a richer set of reward metrics (aligning closely with user preferences), without the need for manual intervention, resulting in visually striking and semantically consistent images.}

As shown in \cref{fig:sp_lexica,fig:sp_diffusiondb}, the generated image $\mathbf{I}^\mathrm{DFP}$ maintains the identity consistency of the image $\mathbf{I}$ produced by the short prompt $\mathbf{s}$ when using the same seed. Meanwhile, it incorporates additional image details that enhance visual appeal. This is evident in \cref{fig:sp_lexica} with the example of a ``commission of a fit male anthro albino lion holding a sword,'' and in \cref{fig:sp_diffusiondb} with ``Grimes with elf ears.'' This capability can be further developed to ensure consistent role generation. To further enhance our model, it is advantageous to incorporate more comprehensive reward considerations. For instance, evaluating generated images based on factors such as high resolution and proportional composition can contribute to their overall quality and realism. Furthermore, to address issues such as attribute leakage and missing objects observed in the original Stable Diffusion method, advanced control techniques can be explored. One potential approach involves incorporating control attention maps into the action space. By selectively directing attention to specific regions in the input image, the model gains finer control over the generation process. Consequently, issues related to attribute leakage can be mitigated, and the preservation of important elements can be ensured. By exploring these possibilities and developing more sophisticated control mechanisms, we can enhance the capabilities of our model and overcome the limitations observed in its current implementation.